\documentclass[12pt]{article}
\usepackage{multirow}
\usepackage{booktabs}
\usepackage{dsfont}
\usepackage{subcaption}
\usepackage{amssymb}
\usepackage{amsmath}
\usepackage{amstext}
\usepackage[authoryear,round]{natbib}
\usepackage{amsthm}
\usepackage{mathrsfs}
\usepackage{graphics} 
\usepackage{color,verbatim}
\usepackage[bookmarks=false,
            colorlinks,
            linkcolor=blue,
            anchorcolor=blue,
            citecolor=blue,
            urlcolor=black
            ]{hyperref}
\usepackage{mathtools}
\usepackage{enumitem}
\usepackage{algorithm}
\usepackage{algpseudocode}
\usepackage{chngcntr}
\usepackage{xr}

\makeatletter
\newcommand*{\addFileDependency}[1]{%
  \typeout{(#1)}%
  \@addtofilelist{#1}
  \IfFileExists{#1}{}{\typeout{No file #1.}}
}
 
\selectfont

\DeclareMathOperator{\sign}{sign}
\DeclareMathOperator{\argmax}{argmax}

\def\squarebox#1{\hbox to #1{\hfill\vbox to #1{\vfill}}}
\def\boxit#1{\vbox{\hrule\hbox{\vrule\kern6pt
      \vbox{\kern6pt#1\kern6pt}\kern6pt\vrule}\hrule}}

\def\boxit#1{\vbox{\hrule\hbox{\vrule\kern6pt
      \vbox{\kern6pt#1\kern6pt}\kern6pt\vrule}\hrule}}

\def\sumk0p{\sum_{k=0}^\ell}

\renewcommand{\tilde}{\widetilde}
\renewcommand{\hat}{\widehat}

\graphicspath{{./Figures/}}
\numberwithin{equation}{section}
\allowdisplaybreaks

\newtheoremstyle{break}% name
  {9pt}%      Space above
  {12pt}%     Space below
  {\upshape}% Body font
  {}%         Indent amount
  {\bfseries}% Thm head font
  {.}%        Punctuation after thm head
  {\newline}% Space after thm head: \newline = linebreak
  {}%         Thm head spec

\newtheorem{rem}{Remark}
\newtheorem{theo}{Theorem} 
\newtheorem{prop}{Proposition}

\newtheorem{defi}{Definition}

\theoremstyle{definition}
\newtheorem{exa}{Example}
\theoremstyle{break}

\theoremstyle{remark}

\usepackage{natbib}
\def\namedlabel#1#2{\begingroup
    #2%
    \def\@currentlabel{#2}%
    \phantomsection\label{#1}\endgroup
} 

\title{}
\author{}
\date{}

\begin{document}

\begin{center}
{\bf \Large 
High-dimensional Semi-supervised Classification via the Fermat Distance} \\[1cm]
\end{center}

\begin{center}
Ruoxu Tan \\
School of Mathematical Sciences and School of Economics and Management, Tongji University,
ruoxut@tongji.edu.cn\\ 
Yiming Zang \\
Department of Sciences,  
North China University of Technology,
yiming.zang@ncut.edu.cn\\
\end{center}
\begin{abstract}
\linespread{1}\selectfont
Semi-supervised classification, where unlabeled data are massive but labeled data are limited, often arises in machine learning applications.
We address this challenge under high-dimensional data by leveraging the manifold and cluster assumptions. Based on the Fermat distance, a density-sensitive metric that naturally encodes the cluster assumption, we propose the weighted $k$-nearest neighbors (NN) classifier and multidimensional scaling (MDS)-induced classifiers. The use of MDS with a large target dimension allows the effective application of linear classifiers to complex manifold data. Theoretically, we derive a sharp lower bound for the expected excess risk within clusters and prove that the weighted $k$-NN classifier utilizing the true Fermat distance is minimax optimal. Furthermore, we explicitly quantify the utility of unlabeled data by showing that the error arising from estimating the Fermat distance decays exponentially with the pooled sample size. Such a rate is much faster than the related rates in the literature. Extensive experiments on synthetic and real datasets demonstrate  competitive or superior performance of our approaches compared to state-of-the-art graph-based semi-supervised classifiers.

\end{abstract}
\textbf{Key words}: clusters; metric learning; minimax rates; multidimensional scaling; optimal risk.

\section{Introduction} 

\linespread{1.7}\selectfont
Classification for high-dimensional data presents a cornerstone challenge in machine learning. We address this challenge under the semi-supervised framework, where labeled data are scarce but unlabeled data are abundant. Such settings are particularly prevalent in medical sciences \citep{Nguyen2023}. For example, while low-dose computed tomography (LDCT) is a standard, rapid tool for early cancer screening, obtaining accurate annotations requires time-consuming evaluation by experts. Consequently, machine learning algorithms that effectively incorporate unlabeled data can provide significant diagnostic improvements. Beyond healthcare, border applications of semi-supervised learning include autonomous driving \citep{Lin2025}, credit risk \citep{Yu2025}, remote sensing \citep{Wang2025}, etc.

However, the mere abundance of unlabeled data does not automatically translate to improved classification performance.
To extract useful information from the unlabeled sample, one must presume connections between the marginal distribution of the covariate and the conditional distribution of the label variable. Our framework is based on the manifold and cluster assumptions: The former assumes that high-dimensional data reside on an unknown low-dimensional manifold, and the latter requires data in a high density region are likely to possess the same label. While both assumptions are ubiquitous in the literature of semi-supervised learning \citep{Rigollet2007,Azizyan2013,Mey2023}, their formalizations and implications vary significantly. In Section~\ref{sc_2}, we will rigorously formalize them, especially the cluster assumption, to establish the theoretical foundation of our model.

Our primary motivation stems from the intrinsic limitations that classical Euclidean and Riemannian metrics do not account for data density. As a result, they fail to exploit the cluster assumption. In contrast, we leverage the power-$\alpha$ Fermat distance with $\alpha>1$ \citep{Hwang2016,Groisman2022,Little2022}, a density-sensitive metric that naturally encodes the cluster assumption: It effectively shortens (Riemannian) distances within high-density clusters and elongates distances across low-density boundaries. Building upon this geometric insight, we propose a set of new classifiers that perform competitively in practice. In theory, we show that the fast convergence rate is achievable with the true Fermat distance and quantify the utility of unlabeled data in estimating the Fermat distance. More specifically, our contributions are summarized as follows:
\begin{itemize}
    \item Under the manifold and cluster assumptions combined with standard regularity assumptions, we derive a lower bound for the expected excess risk within clusters. The lower bound decays exponentially with the labeled sample size, which appears conservative but theoretically attainable.

    \item Built upon the estimated Fermat distance, we propose the weighted $k$-nearest neighbors (NN) classifier and multidimensional scaling (MDS)-induced classifiers. The latter utilizes MDS to generate data representations in a potentially high target dimension, which seems to be a new strategy. The generated data are then seamlessly combined with \emph{any} multivariate classifier. Notably, we show that the linear support vector machines (SVM) with this MDS strategy performs among the best in experiments.

    \item Theoretically, we prove that the weighted $k$-NN classifier with the true Fermat distance is minimax optimal, i.e., the upper bound of its expected excess risk within clusters matches to the lower bound up to a logarithmic term. For the version with the estimated Fermat distance, we show that the additional error term decays exponentially with the pooled (i.e., labeled and unlabeled) sample size. This exponential term represents a much faster convergence rate compared to the related rate in \citet{Rigollet2007}. It follows that, with a sufficiently large unlabeled sample size, the classifier with the estimated Fermat distance performs as if the Fermat distance were known.
\end{itemize}

We demonstrate the efficacy of our proposed classifiers through extensive experiments on synthetic and real data examples, including image and gene datasets that are most likely to satisfy the manifold and cluster assumptions. Our classifiers achieve competitive or superior accuracy compared to state-of-the-art graph-based semi-supervised classifiers \citep{Calder2020,Mai2021,Calder2022}.

The rest of the article is organized as follows. We summarize related work in Section~\ref{sc_related}. In Section~\ref{sc_2}, after introducing our model and assumptions, we derive the lower bound. In Section~\ref{sc_class}, we first show how to estimate the Fermat distance, and then introduce our proposed classifiers. We derive upper bounds of the $k$-NN related classifiers in Section~\ref{sc_theo}. Experiments on both synthetic and real data examples are given in Section~\ref{sc_exper}. We conclude with discussion in Section~\ref{sc_dis}. Proofs and additional experimental results are included in the Supplementary Material.

\subsection{Related Work}\label{sc_related}
The literature on semi-supervised learning is massive with a predominant focus on classification tasks. The seminal works taking advantage of manifold by \citet{Belkin2004} and \citet{Belkin2006} pioneer the field of manifold regularization, which is later referred to as the graph Laplacian or graph-based approaches. Recent advances in this domain include the Poisson learning \citep{Calder2020,Calder2022}, the graph-based centered regularization \citep{Mai2021}, etc. \citet{Wang2025a} proposed a random-projection approach for high-dimensional data, which can be combined with clustering and semi-supervised algorithms. Early theoretical works on classification performance include \citet{Castelli1996} and \citet{Rigollet2007}. Due to its capability of dealing with massive and structured data, approaches based on deep neural networks gain popularity \citep{Sohn2020,Yang2024}. As it is impossible to give a full review, we refer readers to \citet{VanEngelen2020} and \citet{Mey2023} for comprehensive surveys.

Beyond classification, a growing number of recent works focused on statistical estimation and inference under the semi-supervised framework, including linear regression \citep{Cai2020,Azriel2022,Deng2024}, nonparametric regression \citep{Azizyan2013}, mean estimation \citep{Zhang2019}, M-estimation \citep{Song2024}, U-statistics \citep{Kim2025}, and inference \citep{Angelopoulos2023,Zrnic2024}. There is also a branch of works trying to avoid negative semi-supervised learning \citep{Loog2015,Guo2020}. 
  
Our methodology can be categorized as metric learning \citep{Kulis2013}. The use of density-sensitive metrics dates back to early 2000s \citep{Sajama2005}. Until the theoretical work by \citet{Hwang2016}, researchers have begun to realize the capacity of the power-weighted shortest path distance, whose limit is now called the Fermat distance. Applications of the Fermat distance on machine learning have recently received great attention, most of which is on unsupervised learning \citep{Mckenzie2019,Little2020,Fernandez2023,Trillos2024}. In this work, we investigate its application to semi-supervised learning. We note that \citet{Bijral2011} explored a variant of the sample Fermat distance for semi-supervised classification; however, their work predated the rigorous theoretical understanding established by \citet{Hwang2016}. Consequently, their definition and computation of the metric differ from ours. Moreover, we distinguish our work by proposing new classifiers and providing the theoretical guarantees that are absent in earlier works.

\section{Model and Lower Bound}\label{sc_2}

\subsection{Model and Data}\label{sc_model}
Suppose that we observe an independent and identically distributed (i.i.d.)~sample of $\{(X_i,Y_i)\}_{i=1}^{n_\ell}$, where $X_i\in \mathbb{R}^D$ denotes the high-dimensional covariate and $Y_i\in \mathcal{Y}=\{0,1,\ldots,K-1\}$ denotes the label variable of $K$ classes. Let $(X,Y)$ denote a generic pair of variables that has the same distribution as $(X_i,Y_i)$, and let $\mathbb{P}_{Z}$ denote the distribution of a generic variable $Z$. We assume that $X$ lies on an unknown compact Riemannian manifold $\mathcal{M}\subset\mathbb{R}^D$ of the intrinsic dimension $d$, the so-called \emph{manifold assumption}. The distributions $\mathbb{P}_{X|Y=y}$ for different $y$ are concentrated on different regions on $\mathcal{M}$, but we allow their supports to intersect. Such a requirement  is more general than the assumption that individuals having different labels lie on different submanifolds \citep{Vural2018}. The ambient dimension $D$ can be (but not necessarily) much larger than $n_\ell$, representing a high-dimensional setting. The manifold assumption has been widely adopted in the high-dimensional data analysis \citep{Whiteley2026}, and it is reasonable in many data applications \citep{Rawat2017,Nguyen2022}. For example, \citet{Gong2019} showed that a large-scale dataset of face images that vary in pose, illumination, and expression is of intrinsic dimension around 10 to 20. 

In addition to the labeled sample $\{(X_i,Y_i)\}_{i=1}^{n_\ell}$, we also observe an independent abundant unlabeled sample $\{X_i\}_{i=n_\ell+1}^{n_\ell+n_u}$, whose distribution is the same as the marginal distribution $\mathbb{P}_X$ from the labeled sample, i.e., the semi-supervised setting. We use $n=n_\ell+n_u$ to denote the total sample size. 
%Although an extension to the case where these two distributions differ, i.e., a transfer learning setting, is possible, it remains beyond the scope of this work. 
The unlabeled sample size $n_u$ is often much larger than the labeled sample size $n_\ell$, corresponding to common applications where collecting covariate observations is cheap but obtaining accurate labels is expensive and limited. Our goal is to construct a better classifier by incorporating the information of the unlabeled sample. To this end, we utilize a density-sensitive metric that is particularly beneficial under the semi-supervised context. 

Specifically, let $\|\cdot\|_{\mathbb{R}^D}$ denote the Euclidean distance on $\mathbb{R}^D$ and $f_X$ denote the density of $X$.
For any $\alpha\geq 1$ and two points $x_1,x_2$ on $\mathcal{M}$, we define the power-$\alpha$ Fermat distance as
\begin{align}\label{eq_fd}
d_{\mathcal{M},\alpha}(x_1,x_2) = \inf_{\gamma}\bigg[\mu_{\alpha,d}\int_{0}^1 \frac{\|\gamma'(t)\|_{\mathbb{R}^D}}{f_X\{\gamma(t)\}^{\frac{\alpha-1}{d}}} \,dt \bigg]^{1/\alpha}\,,
\end{align}
where $\mu_{\alpha,d}$ is the percolation constant that depends only on $\alpha$ and $d$, and $\gamma:[0,1]\to\mathcal{M}$ is a continuously differentiable path satisfying $\gamma(0)=x_1$ and $\gamma(1)=x_2$. Slightly different from the literature~\citep{Hwang2016,Little2022,Fernandez2023,Trillos2024}, we incorporate the percolation constant $\mu_{\alpha,d}$ in the definition of $d_{\mathcal{M},\alpha}$ so that the notation of theoretical results is relieved. The constant $\mu_{\alpha,d}$ is defined as a limit with $\alpha>1$ \citep{Howard1997}. Letting $d_{\mathcal{M}}$ denote the Riemannian  metric induced from the inner product in $\mathbb{R}^D$, we define $\mu_{1,d}=1$ so that the Fermat distance reduces to the geodesic distance with $\alpha=1$, i.e., $d_{\mathcal{M},1} = d_{\mathcal{M}}$. 

Interesting situations arise when $\alpha>1$ and $f_X$ is not a constant, under which the optimal path $\gamma$ in~\eqref{eq_fd} is generally not the geodesic path but favors regions where the density $f_X$ is high. The Fermat distance with $\alpha>1$ is thus a density-sensitive unsupervised metric: Points belonging to a high-density region tend to have shorter Fermat distances. To see why this property is particularly attractive under the semi-supervised learning, we need to make a proper assumption where the marginal distribution $\mathbb{P}_X$ and the conditional distribution $\mathbb{P}_{Y|X}$ are connected. Following the literature on semi-supervised learning \citep{Rigollet2007,Azizyan2013,Mey2023}, we postulate the \emph{cluster assumption} in our context: For two points $X_1$ and $X_2$ in a region of relatively high density of the manifold, i.e., a cluster, the corresponding labels $Y_1$ and $Y_2$ are likely to be the same. Under the cluster assumption, the unlabeled sample facilitates classification by providing information about the clusters. This assumption is intuitively reasonable and prevailing in the literature. We will discuss this  assumption in more details in Section~\ref{sc_lower}.

The Fermat distance becomes attractive with the cluster assumption. For example, consider three points $x_1,x_2$ and $x_3$ with $d_{\mathcal{M}}(x_1,x_2)=d_{\mathcal{M}}(x_1,x_3)$, i.e., the geodesic distance between $x_1$ and $x_2$ is the same as that between $x_1$ and $x_3$. Classifiers based on the geodesic distance can only draw the same conclusion for $x_2$ and $x_3$. Now suppose that the geodesic path of $x_1$ and $x_2$ passes through a region where $f_X$ is low, while that of $x_1$ and $x_3$ only passes through a high-density region (so that they belong to the same cluster). It follows that $d_{\mathcal{M},\alpha}(x_1,x_2)>d_{\mathcal{M},\alpha}(x_1,x_3)$, for $\alpha>1$. If we define a classifier such that a smaller $d_{\mathcal{M},\alpha}$ indicates a higher probability of possessing the same label, then this classifier properly reflects the cluster assumption, suggesting that $x_1$ and $x_3$ tend to have the same label. In this way, the unlabeled sample is helpful for classification, and we will follow this idea to define classifiers in Section~\ref{sc_class}. 

\subsection{Assumptions and Lower Bound}\label{sc_lower}
Before introducing our classifiers, we discuss more on model assumptions and derive a lower bound. In theory, we focus on binary classification for brevity, i.e., $Y\in\{0,1\}$, while we allow more classes in practice. Classically, a cluster is defined as a connected region where the density is high. Since the power-$\alpha$ Fermat distance induces a path on $\mathcal{M}$, it is quite natural to define a cluster in our context as follows.
\begin{defi}[Cluster]
For fixed $\alpha \geq 1$ and $\kappa>0$, a closed subset $T\subset \mathcal{M}$ is a cluster with respect to $\alpha$ and $\kappa$, if $\forall x_1,x_2\in T$, the optimal path $\gamma_{12}$ found by the power-$\alpha$ Fermat distance $d_{\mathcal{M},\alpha}(x_1,x_2)$ satisfies $\gamma_{12}(t)\in T$ and $f_{X} \{\gamma_{12}(t)\}\geq \kappa$,  $\forall t\in[0,1]$.
\end{defi}

We assume that there is a set of disjoint clusters $\{T_j\}_{j=1}^m$, which mainly serve to explain the theoretical properties of classifiers. Our classifiers introduced in Section~\ref{sc_class} do not need to estimate the clusters. Throughout, we regard $\{T_j\}_{j=1}^m$ as fixed but unknown subsets of $\mathcal{M}$. We explicitly define the cluster assumption as follows. 
\begin{itemize}
\item[\namedlabel{CA1}{(A1)}](Cluster Assumption) There exists a $\epsilon_0>0$ such that $\inf_{x\in\cup_{j=1}^m T_j}|2\eta(x)-1|>\epsilon_0$, where $\eta (x) = P(Y=1|X=x)$. 
\end{itemize}
The cluster assumption is slightly stronger than the one defined in \citet{Rigollet2007}, who assumed that the function $\mathds{1}\{\eta(x)>1/2\}$ is a constant on each cluster and developed a semi-supervised classifier based on estimating the clusters. Here, we require there is a strict gap between $\eta(x)$ and $1/2$ for tractable theoretical investigation. It is always possible to obtain \ref{CA1} by assuming \citet{Rigollet2007}'s cluster assumption and slightly shrinking the clusters. The key implication by~\ref{CA1} is that, for all $T_j$,
\begin{align*}
P(Y_1=Y_2|X_1,X_2\in T_j)\geq P(Y_1\neq Y_2|X_1,X_2\in T_j)\,,
\end{align*}
which directly reflects the role of clusters in task of classification. 

It is worthwhile to compare the cluster assumption~\ref{CA1} with the Tsybakov margin assumption \citep{Mammen1999,Audibert2007} used to show fast convergence rates in classification or discrimination problems : There exists $\gamma>0$ and $C>0$ such that $\forall \epsilon>0$,
\begin{align*}
 P(0<|2\eta(X)-1|\leq \epsilon)\leq C \epsilon^\gamma\,.
\end{align*}
That is, the probability that $\eta$ is close to $1/2$ is polynomial decaying. 
When $\gamma=\infty$, the assumption above implies that there exists $h>0$ such that $\forall x \in\mathrm{supp}(f_X)$,
\begin{align}\label{mar_cond}
    |2\eta(x)-1|>h\,,
\end{align}
which has been invoked in \citet{Massart2006} to prove various minimax bounds for the empirical risk minimization estimators. The assumption~\eqref{mar_cond} is clearly stronger than the cluster assumption \ref{CA1}, because we only require such a gap exists for clusters. Therefore, \ref{CA1} is reasonable under our context, and indeed, it plays a vital role in deriving fast convergence rates. We further illustrate the cluster assumption with an example below.

\begin{figure}[t]
\centering
\includegraphics[width=0.5\textwidth]{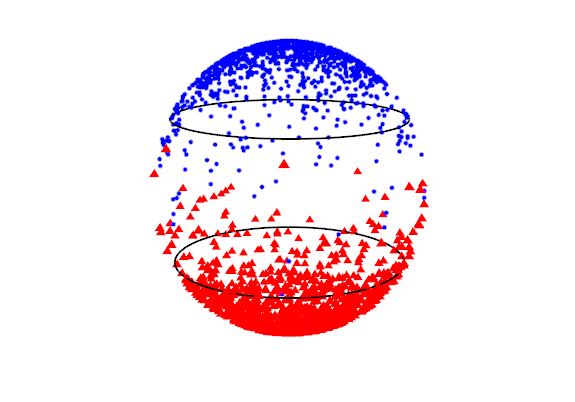}
\caption{A sample from the distribution described in Example~\ref{exa_1}: The circle and triangle points correspond two classes; the two latitude lines are boundaries of two clusters.}\label{fg_exa}
\end{figure}

\begin{exa}[Clusters on a sphere]\label{exa_1}
Let $P(Y=0)=P(Y=1)=1/2$ and $X|Y=y$ follow a von Mises-Fisher distribution $vM(\mu_y,\sigma)$, i.e., $f_{X|Y=y}(x) = C_\sigma\exp(\sigma x^\top \mu_y)$, where $C_\sigma$ is a constant depending on $\sigma$, $\mu_y\in\mathbb{R}^3$ is the mean direction parameter, and $\sigma\geq 0$ is the concentration parameter. A von Mises-Fisher distribution on $\mathbb{S}^{d-1}$ is restriction of a $\mathbb{R}^d$-variate normal distribution to $\mathbb{S}^{d-1}$ up to a rescaling. Now set $\mu_0 = (0,0,1)^\top$, $\mu_1 = (0,0,-1)^\top$, $\alpha = 1$, $\sigma = \sigma_0$ and $\kappa=\kappa_0$, then the two clusters, corresponding to $Y=0$ and $Y=1$, are two surface areas of the polar caps bounded by two lines of latitude; see Figure~\ref{fg_exa} for illustration. For example, we can compute that the boundary lines of latitude at 30° North and 30° South correspond to $\sigma_0=5$ and $\kappa_0 = 0.82$. The cluster assumption~\ref{CA1} always holds for any $\sigma_0$ and $\kappa_0$ such that the boundary latitude lines are not the equator. 
\end{exa}

In addition to the cluster assumption~\ref{CA1}, the following assumptions are also needed to regularize the joint distribution $\mathbb{P}_{X,Y}$.
\begin{itemize}
\item[\namedlabel{CA2}{(A2)}] (Lipschitz Condition) The conditional distribution $\eta(\cdot)$ is Lipschitz continuous with respect to the power-$\alpha$ Fermat distance, i.e., $\exists L>0$ such that $\forall x_1,x_2\in \mathcal{M}$, $|\eta(x_1)-\eta(x_2)|\leq Ld_{\mathcal{M},\alpha}(x_1,x_2)$.

\item[\namedlabel{CA3}{(A3)}] (Minimal Mass Assumption) For all $x\in\mathcal{M}$, there exists $t_0>0$ such that $\forall t\leq t_0$, $P(d_{\mathcal{M},\alpha}(X,x)\leq t)\geq Cf_X(x)t^d$. 
 
\end{itemize}
Assumption~\ref{CA2} is the Lipschitz condition adapted to the Fermat distance, while Assumption~\ref{CA3} is the minimal mass assumption with respect to the Fermat distance. Their Euclidean versions are commonly assumed in the literature on the classification theory; see e.g., \citet{Audibert2007} and \citet{Gadat2016}.

%According to the proposition 3.1 in \citet{Gadat2016}, if the density $f_X$ is bounded below, then $f_X$ is bounded above and $(c_0,r_0)$-regular \citep{Audibert2007}. Therefore, we can deduce that $f_X$ is bounded above and $(c_0,r_0)$-regular adapted to the Riemannian measure if we restrict $f_X$ on the clusters, because $f_X(x)$ is bounded below, for $x\in\cup_{j=1}^m T_j$.  

For a classifier $h:\mathbb{R}^D\to \{0,1\}$, let $\mathcal{R}(h)=P\{Y\neq h(X)\}$ denote the classification risk. Let $h^*(x) = \mathds{1} \{\eta(x)>1/2\}$ denote the Bayes classifier. It is well known that $h^*$ achieves the optimal risk, i.e., $\mathcal{R}(h^*)\leq \mathcal{R}(h)$, for any classifier $h$. It is then of interest to develop classifiers whose risk converges to $\mathcal{R}(h^*)$. However, it has been shown in \citet{Rigollet2007} that the risk outside the clusters cannot be improved using an additional unlabeled sample, essentially because the cluster assumption only concerns data within clusters. Therefore, instead of $\mathcal{R}(h)$, we focus on the risk within the clusters, $\mathcal{R}_{\mathcal{T}}(h) = P(Y\neq h(X),X\in \cup_{j=1}^m T_j)$, for any classifier $h$. Since we will consider both the supervised and the semi-supervised case, we use $E_{n_\ell}(\cdot)$ to denote the expectation with respect to $\{X_i,Y_i\}_{i=1}^{n_\ell}$. Similarly, $E_{n}(\cdot)$ denotes the expectation with respect to the pooled sample $\{X_i,Y_i\}_{i=1}^{n_\ell}\cup\{X_i\}_{i=n_\ell+1}^n$, and $P_{n_\ell}(\cdot)$ and $P_n(\cdot)$ denote the corresponding probabilities. Throughout, we use $C$ to denote a generic positive constant that can differ under different contexts. We denote by $\mathcal{P}(\epsilon_0,L,d)$ the space of joint distributions of $(X,Y)$ satisfying Assumptions~\ref{CA1} to \ref{CA3}. The following theorem shows the lower bound of $E_{n_\ell}\{\mathcal{R}_{\mathcal{T}}(h)\}-\mathcal{R}_{\mathcal{T}}(h^*)$ for distributions in $\mathcal{P}(\epsilon_0,L,d)$. 
\begin{theo}\label{theo_lower}
Under Assumptions~\ref{CA1} to \ref{CA3}, we have, for all $n_\ell\in\mathbb{N}^+$,
\begin{align*}
\inf_{\hat{h}} \sup_{\mathbb{P}_{X,Y}\in\mathcal{P}(\epsilon_0,L,d)}    E_{n_\ell}\{\mathcal{R}_\mathcal{T}(\hat{h})\}-\mathcal{R}_\mathcal{T}(h^*)\geq C_1\exp(-C_2n_\ell)\,,    
\end{align*}
for some constants $C_1,C_2>0$.
\end{theo}
The proof is given in Section~A of the Supplementary Material. This theorem establishes a lower bound of the expected excess risk within clusters that decays exponentially with the labeled sample size $n_\ell$. Since the bound holds for all finite sample size $n_\ell$, it incorporates a constant $C_1$ in front of the exponential term. Such an exponential lower bound may seem conservative, but if we are able to develop classifiers whose upper bounds match the lower bound, then the lower bound is tight. Note that this result represents a supervised lower bound under the standard class of distributions $\mathcal{P}(\epsilon_0,L,d)$. We do not seek to show that our semi-supervised classifiers can achieve a faster convergence rate, as doing so requires a highly subtle construction of the class of distributions of $\mathbb{P}_{X,Y}$ \citep{Azizyan2013}. Instead, our aims are twofold: To develop classifiers based on the true Fermat distance achieving the minimax convergence rate, and to explicitly quantify the utility of unlabeled samples in estimating the Fermat distance.

\section{Semi-supervised Classification}\label{sc_class}

\subsection{Estimation of Fermat Distance}\label{sc_FD}

%To utilize the cluster assumption for semi-supervised classification, one may try to directly estimate the clusters by density level sets~\citep{Rigollet2007}. However, density-based clustering is sensitive to tuning parameters~\citep{Rinaldo2012}, and it is unclear how to classify data outside of clusters. Instead, we exploit the cluster assumption by a density-sensitive metric that is particularly beneficial under the semi-supervised context. 
Since the Fermat distance is unsupervised, we can incorporate both the labeled and unlabeled samples to construct an estimator. 
Let $G$ denote an adjacency graph for the pooled sample $\{X_i\}_{i=1}^n$, which is represented by a $n\times n$ matrix whose $(i,j)$-th entry is 1 if the individuals $i$ and $j$ are adjacent and 0 otherwise. A discrete path $\gamma_m$ in $G$ between $i$ and $j$ is an ordered sequence $\gamma_m=(i=i_1,\ldots,i_m=j)$ such that $G(i_{k},i_{k+1})=1$, for $k = 1,\ldots,m-1$.
To estimate $d_{\mathcal{M},\alpha}$ in a finite sample, the standard approach is based on the complete graph $G_{\rm{com}}$ of  $\{X_i\}_{i=1}^n$,
\begin{align}\label{eq_theo_fd_tilde}
\tilde{d}_{G_\textrm{com},\alpha} (X_i,X_j) = \min_{\gamma_m\in G_{\textrm{com}}}\Big(\sum_{j=1}^m \|X_{i_j}-X_{i_{j+1}}\|_{\mathbb{R}^D}^\alpha\Big)^{1/\alpha}\,.
\end{align} 
When $\alpha=1$, $\tilde{d}_{\textrm{com},\alpha} $ is merely the Euclidean distance in $\mathbb{R}^D$. As $\alpha$ increases, large segment lengths $\|X_{i_j}-X_{i_{j+1}}\|_{\mathbb{R}^D}$ are further enlarged so that the optimal path prefers shorter segments where the density is high. It has been be shown that, for $\alpha>1$, $n^{\frac{\alpha-1}{\alpha d}}\tilde{d}_{\textrm{com},\alpha} (X_i,X_j)\to  d_{\mathcal{M},\alpha} (X_i,X_j)$ 
almost surely as $n\to \infty$ \citep{Hwang2016,Fernandez2023}. Therefore, $d_{\mathcal{M},\alpha}$ can be consistently estimated by $n^{\frac{\alpha-1}{\alpha d}}\tilde{d}_{\textrm{com},\alpha} $, for $\alpha>1$.

However, computing shortest paths on the complete graph $G_{\rm{com}}$ is too heavy for large datasets. A common alternative is finding the optimal path $\gamma_m$ restricted to a relatively sparse graph \citep{Little2022}, such as connecting nearby individuals using $\epsilon$-balls or $k$-nearest neighbors ($k$-NN). However, these methods do not  guarantee the graph connectivity. Note that if a graph has more than one connected component, then the distances between points from different components are not well defined. 
To fix the issue, we propose a simple and effective remedy by taking the union of a $k$-NN graph and the minimum spanning tree graph as the adjacency graph, both of which are based on Euclidean distance in $\mathbb{R}^D$. We denote this graph by $G_{\textrm{$k$-NM}}$, and an alternative to~\eqref{eq_theo_fd_tilde} is $\tilde{d}_{G_\textrm{$k$-NM},\alpha} (X_i,X_j) = \min_{\gamma_m\in G_{\textrm{$k$-NM}}}(\sum_{j=1}^m \|X_{i_j}-X_{i_{j+1}}\|_{\mathbb{R}^D}^\alpha)^{1/\alpha}$. This minimization problem can be effectively solved by the Dijkstra algorithm, a greedy algorithm ensuring the global optimum. It follows that we define the estimator of the Fermat distance by 
\begin{align}\label{eq_fd_est}
\hat{d}_{\mathcal{M},\alpha} (X_i,X_j) = n^{\frac{\alpha-1}{\alpha d}}\tilde{d}_{G,\alpha} (X_i,X_j)\,,
\end{align} 
for all $X_i$ and $X_j$ in the pooled sample $\{X_i\}_{i=1}^n$. Here, $G$ can be $G_{\textrm{com}}$ or $G_{\textrm{$k$-NM}}$. We will show that using $G_{\textrm{com}}$ and using $G_{\textrm{$k$-NM}}$ yield almost the same classification performance, while the latter is computationally much faster than the former when $n=3000$; see Section~\ref{sc_sim} for details.

The procedure above yields the estimated Fermat distance $\hat{d}_{\mathcal{M},\alpha}$ for all pairs in the pooled sample. To compute $\hat{d}_{\mathcal{M},\alpha}(X_{n+1},X_i)$, for $X_i\in\{X_j\}_{j=1}^{n}$ and a future observation $X_{n+1}$, we may update the pooled sample by including $X_{n+1}$ and proceed as above. However, this requires recomputing the whole adjacency graph $G$, which is time-consuming. A simple and effective way to fix this issue is updating the current $G$ by connecting $X_{n+1}$ to its closest $k_0$ points based on the Euclidean distance in $\mathbb{R}^D$. That is, given the current $G$, we update it by including $X_{n+1}$ as a vertex and the $(X_{n+1},X_{(i)})$'s as edges, for $i=1,\ldots,k_0$, where $\|X_{(1)}-X_{n+1}\|_{\mathbb{R}^D}\leq \|X_{(2)}-X_{n+1}\|_{\mathbb{R}^D}\leq \cdots\leq \|X_{(n)}-X_{n+1}\|_{\mathbb{R}^D}$. Here, the value of $k_0$ can be specified as the average degree of $G$. The estimator $\hat{d}_{\mathcal{M},\alpha}(X_{n+1},X_i)$ is then computed based on the updated $G$.

\subsection{Weighted $k$-NN Classifier}\label{sc_wknn}
Now that we have obtained the estimated Fermat distance $\hat{d}_{\mathcal{M},\alpha}$ for any pair of observations, presumably the simplest classification algorithm based on $\hat{d}_{\mathcal{M},\alpha}$ is the $k$-NN classifier. To better incorporate the cluster information contained in the Fermat distance, we propose the weighted $k$-NN classifier based on $\hat{d}_{\mathcal{M},\alpha}$ as follows. 

For an unlabeled observation $X_0$, letting $\{X_{(i)},Y_{(i)}\}_{i=1}^{n_\ell}$ be a reordered sample of $\{X_i,Y_i\}_{i=1}^{n_\ell}$ such that $\hat{d}_{\mathcal{M},\alpha}(X_{(1)},X_0)\leq\cdots \leq \hat{d}_{\mathcal{M},\alpha}(X_{(n_\ell)},X_0)$, we define
the weighted $k$-NN classifier $\hat{h}_{\textrm{w$k$NN}}$ by
\begin{align}\label{wknn}
\hat{h}_{\textrm{w$k$NN}}(X_0)=\hat{Y}  = \underset{y\in\mathcal{Y}}{\argmax}\sum_{i=1}^k\hat{w}_i\mathds{1}\{Y_{(i)}=y\}\,,
\end{align}
where
\begin{align*}
\hat{w}_i = \frac{\exp\{-\hat{d}_{\mathcal{M},\alpha}(X_{(i)},X_0)/\sigma\}}{\sum_{j=1}^k\exp\{-\hat{d}_{\mathcal{M},\alpha}(X_{(j)},X_0)/\sigma\}}\,.
\end{align*}
Here, $\sigma>0$ is a tuning parameter that can be chosen via, for example, the cross validation (CV) procedure with the classification accuracy on the labeled sample. The unweighted $k$-NN classifier is recovered if $\sigma\to \infty$. The weight $\hat{w}_i$ is negatively related to the distance $\hat{d}_{\mathcal{M},\alpha}(X_{(i)},X_0)$, and thus more weights are placed on those with smaller estimated Fermat distances, indicating that they possess the same label by the cluster assumption.
%Therefore, the weighted $k$-NN classifier is generally more effective than the unweighted one, which is validated through simulations in Section~\ref{sc_sim}.

As indicated in our theory (Theorems~\ref{theo_1} and \ref{theo_3} in Section~\ref{sc_theo}), the optimal $k\propto n_\ell/\log(n_\ell) $, which is larger than the classical supervised setting \citep{Gadat2016}. In practice, we suggest setting $k \in (n_\ell /(2K),n_\ell/K)$, where $K$ is the number of classes, for a class-balanced sample. 
 
\subsection{MDS-induced Classifiers}\label{sc_MDS}
It is also possible to define other classification algorithms based on $\hat{d}_{\mathcal{M},\alpha}$, as long as the chosen algorithm only requires pairwise distances. For example, the support vector machines (SVM) with the radial basis function (RBF) kernels \citep{Schoelkopf2002} can be combined with $d_{\mathcal{M},\alpha}$.

In fact, \emph{any} multivariate classifier can be built upon $\hat{d}_{\mathcal{M},\alpha}$ by using the multidimensional scaling \citep[MDS,][]{Borg2007}. Indeed, for a target dimension $p\in \mathbb{N}^+$, we can apply MDS on the estimated distance matrix $\mathbb{D}_\alpha=\big(\hat{d}_{\mathcal{M},\alpha}(X_i,X_j)\big)_{1\leq i,j\leq n}$ to obtain data representations $\{X_i^{\rm{MDS}}\}_{i=1}^{n}\subset \mathbb{R}^p$, whose pairwise Euclidean distances approximate the estimated Fermat distances. A generic multivariate classifier $\tilde{h}$ can then be trained on $\{X_i^{\rm{MDS}},Y_i\}_{i=1}^{n_\ell}$, and, for an unlabeled observation $X_j\in\{X_i\}_{i=n_\ell+1}^n$, we predict its label $Y_j$ by $\tilde{h}(X_j^{\rm{MDS}})$.  

While MDS is conventionally used for dimension reduction \citep{Tenenbaum2000}, we focus on its original use, i.e., metric preservation. Because choosing a small $p$ necessarily distorts the structure of the distance matrix $\mathbb{D}_{\alpha}$, we suggest selecting a sufficiently large $p$ (allowed to be larger than $D$) so that $\|X_i^{\rm{MDS}}-X_j^{\rm{MDS}}\|_{\mathbb{R}^p}$ is arbitrarily close to $\hat{d}_{\mathcal{M},\alpha}(X_i,X_j)$. In fact, a perfect (discrete) isometric embedding, i.e., $\|X_i^{\rm{MDS}}-X_j^{\rm{MDS}}\|_{\mathbb{R}^p}=\hat{d}_{\mathcal{M},\alpha}(X_i,X_j),\forall i \textrm{ and } j$, is guaranteed when $p=n-2$ \citep{Borg2007}. We essentially linearize the manifold structure via the isometric embedding, allowing standard linear classifiers to perform well on these high-dimensional representations $\{X_i^{\rm{MDS}},Y_i\}_{i=1}^{n_\ell}$. For example, we will use this strategy in our numerical experiments to perform the linear SVM based on $\hat{d}_{\mathcal{M},\alpha}$. In particular, we will show by synthetic examples that the linear SVM with MDS preserving distances (i.e., $p$ as large as needed) performs better than that with  MDS reducing dimension (i.e., $p=d$); see Section~\ref{sc_sim} for details.

Although our MDS-induced classifiers are sufficiently inclusive, we suggest restricting to the classifiers that exploit distance-based information such as the $k$-NN and SVM; because otherwise, the benefit of using the Fermat distance is unclear.

\section{Upper Bounds}\label{sc_theo}

In this section, we derive upper bounds of the expected excess risk within clusters for our classifiers. Since the MDS-induced classifiers depend on a particular chosen classifier, we focus on the weighted $k$-NN classifier. Recall that we focus on binary classification in theory.
We first consider the weighted $k$-NN classifier $\tilde{h}_{\textrm{w$k$NN}}$ based on the true power-$\alpha$ Fermat distance: $\tilde{h}_{\textrm{w$k$NN}}(x) =1$ if $\sum_{j=1}^k \tilde{w}_jY_{\pi(j)}(x)>1/2$ and $\tilde{h}_{\textrm{w$k$NN}}(x) =0$ otherwise,  where $\tilde{w}_j=\exp\{- d_{\mathcal{M},\alpha}(X_{\pi(j)},x)/\sigma\}/\sum_{j'=1}^k \exp\{-d_{\mathcal{M},\alpha}(X_{\pi(j')},x)/\sigma\}$, and $\{X_{\pi(i)},Y_{\pi(i)}\}_{i=1}^{n_\ell}$ is a reordered sample of $\{X_i,Y_i\}_{i=1}^{n_\ell}$ such that $d_{\mathcal{M},\alpha}(X_{\pi(1)},x)\leq \ldots\leq d_{\mathcal{M},\alpha}(X_{\pi(n_\ell)},x)$. Note that $\{X_{\pi(i)},Y_{\pi(i)}\}_{i=1}^{n_\ell}$ depend on $x$, while the notation is simplified. For two sequences $a_n$ and $b_n$, we write $a_n\asymp b_n$, if there exists a constant $C>1$ such that $C^{-1}a_n\leq b_n\leq C a_n$, uniformly for all $n\in\mathbb{N}^+$. 
\begin{theo}\label{theo_1}
Under Assumptions~\ref{CA1} to \ref{CA3}, if $k \asymp \lfloor n_\ell/\log(n_\ell) \rfloor$, and $\sigma\asymp (k/n_\ell)^{1/d}$, then there exists a constant $C>0$ and $N_{\ell0}\in\mathbb{N}^+$ such that for all $n_\ell>N_{\ell0}$,
\begin{align*}
\sup_{\mathbb{P}_{X,Y}\in\mathcal{P}(\epsilon_0,L,d)}E_{n_\ell}\{\mathcal{R}_{\mathcal{T}}(\tilde{h}_{\textrm{w$k$NN}})\}-\mathcal{R}_{\mathcal{T}}(h^*)\leq  \exp\Big\{- C\frac{n_\ell}{\log(n_\ell)} \Big\}\,.
\end{align*}
\end{theo} 

The proof of Theorem~\ref{theo_1} is given in Section~B of the Supplementary Material. Compared to the lower bound in Theorem~\ref{theo_lower}, we conclude that  $\tilde{h}_{\textrm{w$k$NN}}$ achieves the minimax rate (up to a logarithm term) with respect to the expected excess risk within the clusters. Notably, the choice of $k$ here is proportional to $n_\ell/\log(n_\ell)$, which is larger than those in the literature \citep[e.g.,][]{Gadat2016}. Intuitively, this deviation is because the cluster assumption~\ref{CA1} inherently suppresses the bias, allowing us to use a larger $k$ to shrink the variance. 

Note that the oracle weighted $k$-NN classifier $\tilde{h}_{\textrm{w$k$NN}}$ does not depend on the clusters.
However, it does depend on the unknown true $d_{\mathcal{M},\alpha}$. Therefore, we proceed to analyze $\hat{h}_{\textrm{w$k$NN}}$ based on the estimated Fermat distance $\hat{d}_{\mathcal{M},\alpha}$, where the unlabeled sample plays a critical role. Recall from~\eqref{eq_fd_est} that $\hat{d}_{\mathcal{M},\alpha}=n^{\frac{\alpha-1}{\alpha d}}\tilde{d}_{G,\alpha}$ with an adjacency graph $G$. We focus on $G=G_{\textrm{com}}$ in theory and leave other choices for future study. The following result is a corollary of Proposition 5 in \citet{Fernandez2023}.

\begin{prop}\label{prop_1}
Suppose that  $f_X$ is smooth and $\inf_{x\in\mathcal{M}} f_X(x)>0$, $\forall \alpha>1$ and  $\forall \epsilon>0$, there exists $\theta_\epsilon>0$ and $N_\epsilon\in\mathbb{N}^+$ such that $\forall n >N_\epsilon$,
\begin{align*}
P_n\Big(\max_{i=1,\ldots,n_\ell}|\hat{d}_{\mathcal{M},\alpha}(X_i,x)-d_{\mathcal{M},\alpha}(X_i,x) |>\epsilon \Big)\leq  \exp \Big(-\theta_\epsilon n^{\frac{1-\lambda }{d+2\alpha}}\Big) \,,
\end{align*}
where $\lambda\in\big((\alpha -1)/\alpha ,1 \big)$ and $n=n_\ell+n_u$ is the pooled sample size.
\end{prop}
The proof of Proposition~\ref{prop_1} is given in Section~C of the Supplementary Material. It shows that for sufficiently large $n$, the tail probability of the event $\max_{i=1,\ldots,n_\ell}|\hat{d}_{\mathcal{M},\alpha}(X_i,x)-d_{\mathcal{M},\alpha}(X_i,x)|$ is exponentially decaying. Here, the unlabeled sample contributes to estimation of the Fermat distance, and thus this probability is with respect to the pooled sample. Although we only consider risk within clusters where the density of $X$ is high, the assumption $\inf_{x\in\mathcal{M}} f_X(x)>0$ is standard in the theory concerning convergence of the Fermat distance \citep{Hwang2016}; otherwise, the Fermat distance may not be well defined for some pair of $(x_1,x_2)$. 

With help of Proposition~\ref{prop_1}, we are able to provide the convergence rate of the expected excess risk within the clusters of $\hat{h}_{\textrm{w$k$NN}}$.
\begin{theo}\label{theo_3}
Under Assumptions~\ref{CA1} to \ref{CA3}, and assume that $f_X$ is smooth with $\inf_{x\in\mathcal{M}} f_X(x)>0$, for all $\alpha>1$, set  $k \asymp \lfloor n_\ell/\log(n_\ell) \rfloor$, $\sigma\asymp (k/n_\ell)^{1/d}$ and $\delta \asymp(k/n_\ell)^{2/d}$, then there exist constants $C>0$,  $\theta_\delta>0$, $N_{\ell0}\in\mathbb{N}^+$ and $N_\delta\in\mathbb{N}^+$ such that $\forall n_\ell>N_{\ell0}$ and
$\forall n =n_\ell+n_u>N_\delta$,  
\begin{align*} 
\sup_{\mathbb{P}_{X,Y}\in\mathcal{P}(\epsilon_0,L,d)}E_{n}\{\mathcal{R}_\mathcal{T}(\hat{h}_{\textrm{w$k$NN}})\}-\mathcal{R}_{\mathcal{T}}(h^*)\leq \exp\Big\{- C\frac{n_\ell}{\log(n_\ell)} \Big\}+\exp\Big(-\theta_\delta n^{\frac{1-\lambda }{d+2\alpha}}\Big) \,,
\end{align*}
where $\lambda\in\big((\alpha -1)/\alpha, 1 \big)$.
\end{theo}
The proof of Theorem~\ref{theo_3} is given in Section~D of the Supplementary Material. Compared to Theorem~\ref{theo_1}, the convergence rate of $\hat{h}_{\textrm{w$k$NN}}$ has the additional term $ \exp(-\theta_\delta n^{\frac{1-\lambda }{d+2\alpha}})$ arising from estimating the Fermat distance. This term explicitly depends on the pooled sample size $n=n_\ell+n_u$. In common applications where $n_u\gg n_\ell$, it is negligible compared to the main term $\exp\{- C n_\ell/\log(n_\ell) \}$, showing the benefit by using the unlabeled sample. That is, provided the unlabeled sample size $n_u$ is sufficiently large, our $\hat{h}_{\textrm{w$k$NN}}$ effectively enjoys the oracle minimax convergence rate as if the true Fermat distance were known. Compared to the related results, this additional exponential term decays much faster than the polynomial term in Theorem~4.1 of \citet{Rigollet2007}, which is due to estimating density level sets. Note that the rate improvement stems from distinct classification approaches rather than the slightly different cluster assumptions.

\section{Experiments}\label{sc_exper}
We compare our Fermat-distance-based (FD-based) classifiers with several state-of-the-art graph-based semi-supervised classifiers in experiments. Specifically, the compared methods include: the Poisson MBO \citep[PMBO,][]{Calder2020} based on resolution of the Poisson equations and a graph-cut enhancement; the Graph-Based Centered Regularization \citep[GBCR,][]{Mai2021} that modifies the Laplacian regularization; the $p$-EIKonal equation \citep[$p$EIK,][]{Calder2022} that yields pairwise distances robust to graph perturbations. PMBO and $p$-EIK are implemented using the Python package \texttt{graphlearning} with $k_p=[n^{1/2}/2]$, the number of NN there, where $[\cdot]$ denotes rounding to the nearest integer. GBCR contains a tuning parameter that is selected by the 5-fold CV with the misclassification risk. We also include naive-$k$NN, the vanilla $k$-NN with Euclidean distance applied on the labeled sample, as a supervised baseline. As for our FD-based classifiers, we consider the weighted $k$-NN classifier (FD-w$k$NN) as in Section~\ref{sc_wknn} and the SVM classifier with the linear kernel (FD-SVM) by utilizing MDS with a high target dimension as in Section~\ref{sc_MDS}. In FD-w$k$NN, the value of $k$ is set $[n_\ell/(1.5K)]$ and the weight $w$ is selected via the 5-fold CV, where $K$ is the number of classes. We fix $\alpha=4$ and $k=[n^{1/2}/2]$ in the adjacency graph $G_{\textrm{$k$-NM}}$ for now. Other choices of $\alpha$, the adjacency graph, and the kernel in FD-SVM will be considered later.
 
\subsection{Synthetic Examples}\label{sc_sim}
We consider a two-moon-shaped data setting that is commonly used for the illustration of nonlinear classification: given $Y_i=0$, $\phi_i = \pi U_i$ and $\theta_i = (\pi-1)\sin(\phi_i)+V_i$; given $Y_i=1$, $\phi_i = \pi U_i+\pi/2$ and $\theta_i = -(\pi-1)\sin(\phi_i-\pi/2)+0.8\pi-V_i$, where $U_i,V_i\overset{iid}{\sim} N(0.5,0.2^2)$. A sample of $\{\phi_i,\theta_i\}_{i=1}^n$ is two-moon-shaped; see Figure S1 in Section E of the Supplementary Material for an illustration. Recall that $n=n_\ell+n_u$ denotes the total sample size. We then lift $\{\phi_i,\theta_i\}_{i=1}^n$ into higher dimensional spaces via the following three models:
\begin{itemize}
    \item[(i)] $X_i = (X_{i1},X_{i2},X_{i3})^\top$, where $X_{i1}= \sin(\theta_i)\cos(\phi_i)$, $X_{i2}=\sin(\theta_i)\sin(\phi_i)$ and $X_{i3}=\cos(\theta_i)$.
    \item[(ii)] $X_i=(X_{i1},\ldots,X_{i500})^\top$, where $X_{ij} = \phi_it_j^2+\theta_i\sin(t_j)$ with $(t_1,\ldots,t_{500})$ an equidistant sequence on $[0,1]$.
    \item[(iii)] $X_i=(X_{i1},\ldots,X_{i500})^\top$, where $X_{i1} = \phi_i+Z_1$, $X_{i2}=\theta_i+Z_2$ and $X_{ij}=Z_j$, for $j=3,\ldots,500$. Here, $Z_i\overset{iid}{\sim}N(0,0.01^2)$.
\end{itemize}

\begin{figure}[t]
\centering 
\includegraphics[width=0.45\textwidth]{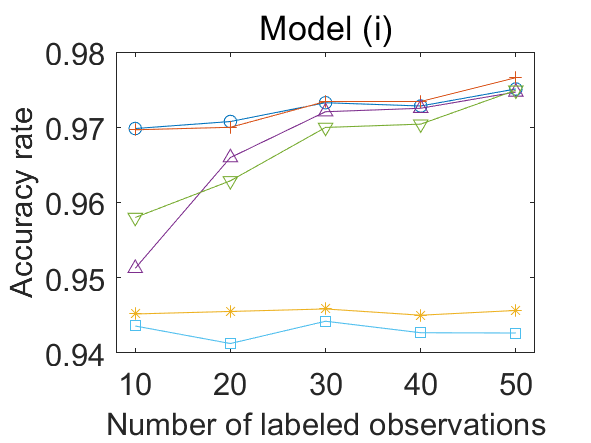}
\includegraphics[width=0.45\textwidth]{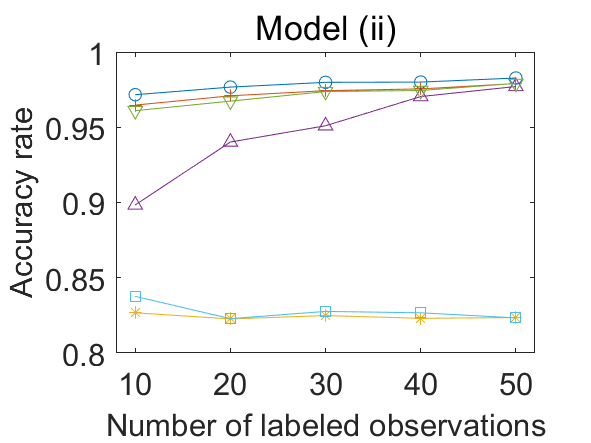}\\
\includegraphics[width=0.45\textwidth]{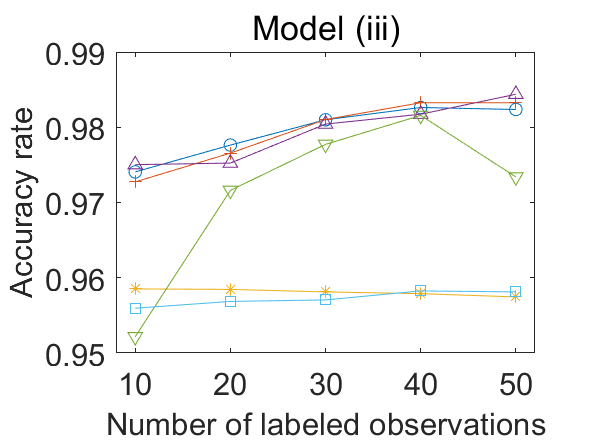}
\includegraphics[width=0.2\textwidth]{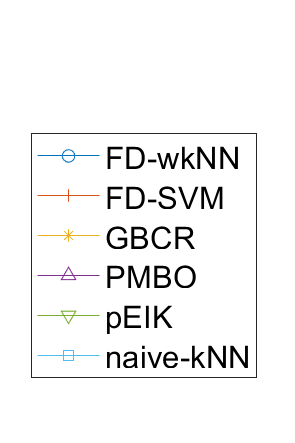}
\caption{The average classification accuracy over 100 repetitions of the six classifiers for the three simulated models. }\label{fg_1}
\end{figure}

Model (i) is a sphere in $\mathbb{R}^3$; model (ii) embed $\{\phi_i,\theta_i\}_{i=1}^n$ into a high-dimensional space by a nonlinear mapping; while model (iii) lift $\{\phi_i,\theta_i\}_{i=1}^n$ to a high-dimensional space by adding a small noise on each coordinate. We fix $n_u=300$ and consider different values in $[10,50]$ for $n_\ell$. In both labeled and unlabeled samples, we generate half of the data with $Y_i=0$ and the other half with $Y_i=1$. Figure~\ref{fg_1} shows the average classification accuracy of the unlabeled data over 100 repetitions for all classifiers.

From Figure~\ref{fg_1}, we see that, under all three models, FD-w$k$NN and FD-SVM perform among the best. $p$EIK (resp., PMBO) performs closely to the FD-based classifiers under model (ii) (resp., model (iii)) but less well under other two models. GBCR performs poorly under our models, possibly because their method is designed for intrinsically high-dimensional data, but the data here are intrinsically low-dimensional. Naive-$k$NN provides a supervised baseline, whose classification accuracy is roughly 3\%-12\% lower than that of our FD-based classifiers under considered models. 

\begin{figure}[t]
\centering
\includegraphics[width=0.32\textwidth]{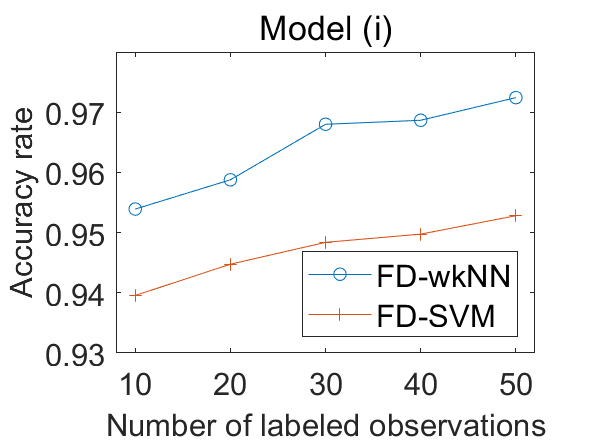}
\includegraphics[width=0.32\textwidth]{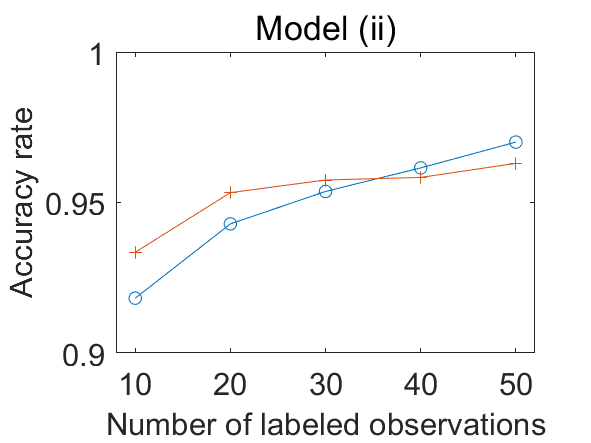}
\includegraphics[width=0.32\textwidth]{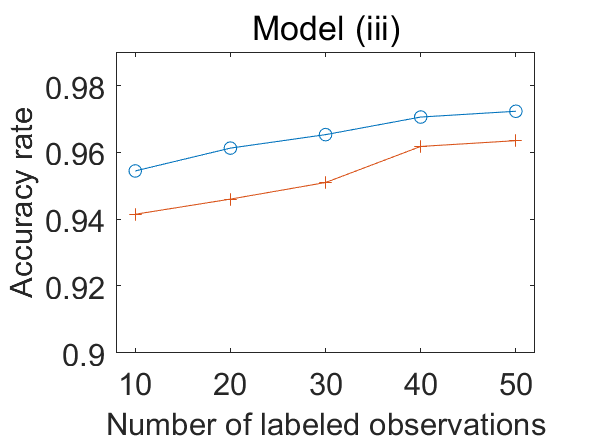}\\ 
\includegraphics[width=0.32\textwidth]{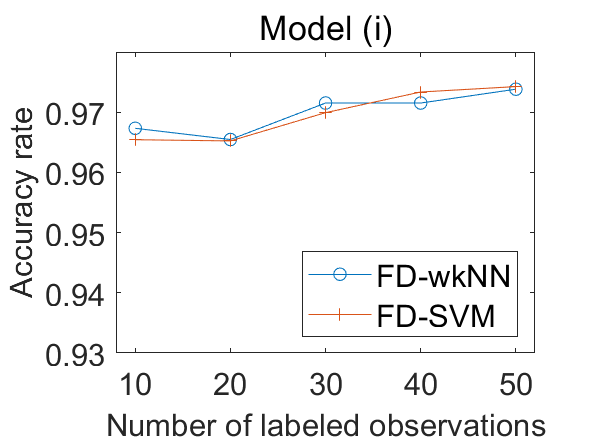}
\includegraphics[width=0.32\textwidth]{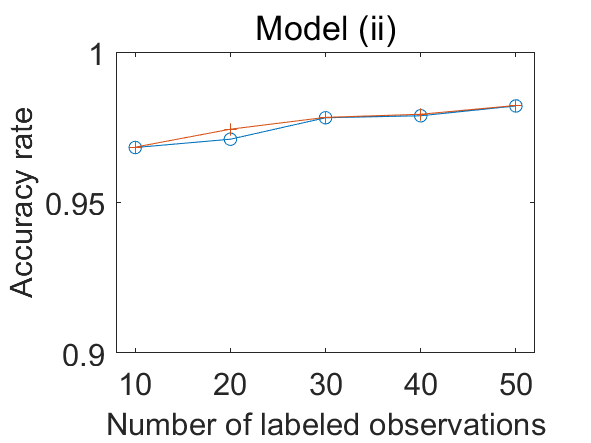}
\includegraphics[width=0.32\textwidth]{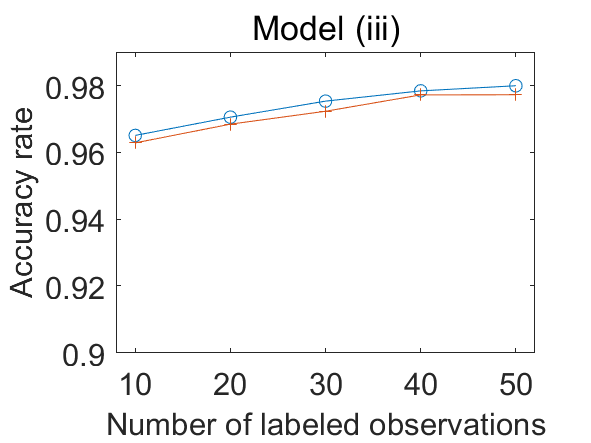}\\ 
\caption{The average classification accuracy over 100 repetitions of the FD-based classifieres with $\alpha=1$ (the first row) and $\alpha=2$ (the second row) for the three simulated models. }\label{fg_2}
\end{figure}

Next, we test a few other values for $\alpha$ in the Fermat distance. In particular, Figure~\ref{fg_2} shows averaged classification performances over 100 repetitions of the FD-based classifiers with $\alpha=1$ and $2$. Figure S2 in Section~E of the Supplementary Material shows those with $\alpha=4 $ and $\alpha=8$. Recall that for $\alpha=1$ and $G=G_{\textrm{$k$-NM}}$, the estimated Fermat distance reduces to the estimated geodesic distance that is independent of the density of covariate. We see from Figures~\ref{fg_2} and S2 that the FD-based classifiers with $\alpha=1$ perform the worst, confirming the effectiveness of using a density-weighted distance. For $\alpha=2,4$ and $8$, the FD-based classifiers perform similarly under our simulated models, suggesting the robustness of our methodology for a small range of $\alpha>1$.

\begin{comment}
Next, we illustrate the benefit by incorporating the unlabeled sample and the weight used in FD-w$k$NN. Specifically, we consider two more classifiers: FD-$k$NN, the version of FD-w$k$NN using the same weight for all individuals; naive-$k$NN, the vanilla $k$-NN based on the Euclidean distance using the labeled sample only. We set $\alpha=4$ and $k=5$ for all these three classifiers. Similar to previous figures, Figure~\ref{fg_3} shows the average classification accuracy over 100 repetitions of the $k$NN-related methods under all models. We see that naive-$k$NN always performs the worst, showing the benefit by incorporating the unlabeled sample using the density-sensitive metric. FD-w$k$NN performs better than FD-$k$NN under models (i) and (iii), indicating usefulness of the weight. 

\begin{figure}[t]
\centering
\includegraphics[width=0.32\textwidth]{model_1_a4_knn_com.png}
\includegraphics[width=0.32\textwidth]{model_2_a4_knn_com.png}
\includegraphics[width=0.32\textwidth]{model_3_a4_knn_com.png}\\  
\caption{The average classification accuracy over 100 repetitions of the $k$NN-related classifiers for the three simulated models. }\label{fg_3}
\end{figure}
\end{comment}

To further illustrate the MDS-induced classifiers, we also test the SVM with the Gaussian kernel, denoted by FD-GSVM, and the version of FD-SVM with the target dimension as the intrinsic dimension, i.e., $p=d$, used in MDS, denoted by FD-$d$SVM. Recall from Section~\ref{sc_MDS} that we aim to maintain the pairwise-estimated Fermat distance as well as possible by using MDS, and thus the target dimension $p$ is chosen as large as needed in FD-SVM. The average classification accuracy of these three SVM-related classifiers is shown in Figure S3 in Section E of the Supplementary Material. We see that FD-SVM performs slightly better than FD-$d$SVM and much better than FD-GSVM. The former shows that our use of MDS as a tool for preserving distance is more useful in terms of classification than that for dimension reduction. The latter is possibly because the target dimension of  MDS is already high, and the SVM with the Gaussian kernel essentially maps the data representations to an infinite-dimensional space, which turns to be harmful for classification.

Finally, we report the computation complexity of our approach and conduct a  running-time comparison with other approaches. Recall from \eqref{eq_fd_est} that estimation of the Fermat distance can be based on the complete graph $G_{\textrm{com}}$ or the sparse graph $G_{\textrm{$k$-NM}}$. We show in Figure S4 in Section E of the Supplementary Material that these two choices result in little difference in terms of classification performance; while using $G_{\textrm{com}}$ is notably slower than using $G_{\textrm{$k$-NM}}$ when $n$ reaches $3000$ as shown in Table S1. Theoretically, the computational complexities of using $G_{\textrm{com}}$ and $G_{\textrm{$k$-NM}}$ are  $O(n^3\log(n))$ and  $O(kn^2\log(n))$, respectively. A running-time comparison is reported in Section E of the Supplementary Material. We see that a FD-based classifier using $G_{\textrm{$k$-NM}}$ only requires around 12 seconds applied on a sample of size $3000$.

\subsection{Real Data Examples}
We further apply our FD-based classifiers and the compared approaches to several real-world datasets. All the parameters are set as stated at the beginning of Section~\ref{sc_exper}. According to the literature on manifold learning with applications \citep{Tenenbaum2000,Belkin2004,Nguyen2022,Yao2024}, image and gene datasets are typical high-dimensional examples that are likely to concentrate on low dimensional manifolds, i.e., satisfying the manifold assumption. Therefore, we consider several such datasets where classification is of primary interest. The first two are image datasets.
\begin{itemize}
    \item (COIL-20) The COIL-20 \citep{Nene1996} is a classical image dataset consisting of 20 objects viewed from different angles. Each object has 72 grayscale images of $128\times 128$ pixels, and thus the total sample size is 1440 with each image regarded as a vector of length 16384.

    \item (f-MNIST) The fashion MNIST dataset \citep{Xiao2017} contains a large number of images of clothing and accessories. Same as the original MNIST dataset, each image is of $28\times 28$ pixels regarded as a 784-dim vector, and the label contains 10 classes. However, the f-MNIST dataset is known to present a more challenging classification task than the original MNIST dataset. Following the literature on manifold learning that utilizes the MNIST dataset \citep{Tenenbaum2000,Belkin2003,Maaten2009}, we randomly select 2000 images from the training sample as our data for computational issue.
\end{itemize}

In addition, we consider two RNA sequence and phenotype datasets from The Cancer Genome Atlas (TCGA), which is a large-scale medical program aimed to catalogue genomic information in different types of cancers. Our aim is to classify the subtypes within a given cancer from RNA expression data. This is a more challenging task than classifying normal and abnormal samples or different types of cancer through gene expression data. We download the following datasets from the UCSC Xena (\url{https://xena.ucsc.edu}), a user-friendly platform for visualizing and analyzing the TCGA data. 
\begin{itemize}
    \item (TCGA-BRCA) The PAM50 subtype of Breast Cancer (BRCA) is famous in cancer categories. In addition to the normal sample, the TCGA-BRCA dataset contains four subtypes of tumor, namely, Luminal A, Luminal B, HER2-enriched and Basal-like. Therefore, the number of classes is 5 with a total sample size of 956. The covariate is $\log_2(\cdot+1)$-transformed gene expression counts of RNA sequence of length 20241, after removing the constant columns.

    \item (TCGA-LUCA) Lung Adenocarcinoma (LUAD) and Lung Squamous Cell Carcinoma (LUSC) are the two predominant subtypes of lung cancer (LUCA). The TCGA-LUCA dataset consists of 576 individuals with the LUAD subtype and 553 individuals with the LUSC subtype. The covariate is the same as that from the TCGA-BRCA dataset.
\end{itemize}

We use the minimal neighborhood method proposed by \citet{Facco2017} to estimate the intrinsic dimension $d$. The estimated $d$ are 8, 14, 40, and 34 for the datasets COIL-20, f-MNIST, TCGA-BRCA, and TCGA-LUCA, respectively, although their original covariate dimensions are much larger. Following the literature on semi-supervised learning \citep[e.g.,][]{Calder2020,Calder2022}, we randomly select a few individuals as the labeled sample and regard the rest as the unlabeled sample. Each of the aforementioned semi-supervised classifiers are applied on the pooled sample (while naive-$k$NN is trained on the labeled sample) and the classification accuracy is computed on the unlabeled sample. The number of labeled individuals per label is set to 4, 8 and 12. The average classification accuracy computed from 100 random selections of the labeled individuals for all classifiers is reported in Table~\ref{tb_real}.

\begin{table}[t]
	\caption{The average classification accuracy $\times100$ (standard deviation $\times100$) of the real data examples computed from 100 random samples with the best result highlight in boldface. }\label{tb_real}
	\centering 
\resizebox{1\columnwidth}{!}{
\begin{tabular}{cccccccc}
\hline
Datasets & $n_\ell/K$& FD-w$k$NN & FD-SVM & GBCR  & PMBO & $p$-EIK & naive-$k$NN\\
\hline
\multirow{3}{*}{COIL-20} & 4 & 88.8 (1.5) & \textbf{91.0} (1.9) & 71.5 (2.4) &  89.5 (1.2) & - & 71.8 (2.3)\\
  & 8 & 92.1 (1.0) & \textbf{95.4} (1.4) & 74.9 (2.4) &  92.2 (1.2) & -  &76.3 (1.6)\\
  & 12 & 94.1 (0.9) & \textbf{97.8} (1.0) & 76.9 (2.8) &  94.2 (1.1) & - &77.7 (1.8)\\
\hline
\multirow{3}{*}{f-MNIST} & 4 & 57.0 (3.4) & 61.3 (3.6) & 55.4 (4.0)  & \textbf{67.5} (2.7) & 65.0 (3.3) & 52.1 (3.8) \\
  & 8 &  61.5 (2.4) & 67.1 (1.9) & 59.3 (2.8)  & \textbf{70.1} (1.5) & 68.7 (2.0)   & 59.8 (2.3) \\
  & 12 &  64.7 (2.2) & 70.1 (1.6) & 60.3 (2.7) & \textbf{71.7} (1.4) & 71.0 (1.5)  & 62.2 (2.0) \\
\hline
\multirow{3}{*}{TCGA-BRCA} & 4 & 70.1 (6.9) & 74.5 (6.5) & 61.0 (6.1)  & \textbf{78.6} (3.5) & 77.7 (3.1) & 66.8 (7.9)\\
& 8 & 75.0 (3.5) & \textbf{79.6} (2.3) & 62.1 (4.5)  & 79.5 (1.8) & 79.3 (2.0) & 73.7 (3.7)\\
 & 12 & 76.8 (3.2) & \textbf{80.7} (1.9) & 62.5 (3.5)  & 80.4 (1.3) & 80.4 (1.4) & 76.4 (3.3)\\
\hline
\multirow{3}{*}{TCGA-LUCA} & 4 & 86.4 (4.2) & \textbf{88.6} (2.4) & 87.0 (1.9)   & 87.3 (1.4) & 87.3 (1.8) & 85.4 (6.0) \\
 & 8 & 88.0 (2.2) & \textbf{89.4} (1.4) & 87.8 (1.8)   & 88.0 (1.6) & 88.2 (1.4) & 88.1 (2.5) \\
 & 12 & 88.9 (1.7) & \textbf{90.0} (1.1) & 88.4 (1.8)   & 88.3 (1.7) & 88.5 (1.5) & 89.0 (1.8)\\

\hline
\end{tabular}
}
\end{table}

Table~\ref{tb_real} shows that our FD-based classifiers, especially FD-SVM, perform competitively among the state-of-the-art approaches. In particular, FD-SVM achieves the highest accuracy in 8 out of the 12 settings. Even when it is not the top performer, it remains highly competitive, e.g., on the f-MNIST dataset. FD-w$k$NN performs less well for the f-MNIST and TCGA-BRCA datasets, but it performs closely to the best approach for the other two datasets. The compared classifiers show varied performance depending on the dataset: PMBO turns to be a strong competitor, achieving outstanding performance on the f-MNIST dataset; $p$-EIK performs similarly to PMBO, except that on the COIL-20 dataset, the average accuracy is unreasonably low and thus not shown; GBCR performs less well overall, possibly because it is designed for intrinsically high-dimensional data; naive-$k$NN performs notably worse than the best semi-supervised classifier for the COIL-20 and f-MNIST datasets, indicating the benefit by incorporating unlabeled samples for image datasets.
 
\section{Discussion}\label{sc_dis}
In this work, we propose the weighted $k$-NN and MDS-induced classifiers based on the Fermat distance, which are designed for high-dimensional semi-supervised classification. We show that the weighted $k$-NN classifier using the true Fermat distance is minimax optimal with respect to the expected excess risk within clusters, and the error due to estimating the Fermat distance decays exponentially with the pooled sample size, which rigorously quantifying the value of unlabeled data. Extensive experiments validate the efficacy of our methodology.

It is of interest to develop upper bounds for some specific MDS-induced classifiers, e.g., the SVM. However, deriving sharp upper bounds for the SVM is non-trivial, as the hinge loss depends on the global data distribution including low-density regions where the cluster assumption provides no regularization. As a result, it remains theoretically unclear how the FD-SVM benefits from the cluster assumption. In addition, extending the framework here to semi-supervised transfer learning \citep{Fan2025} or investigating the use of Fermat distance in other estimation or inference problems \citep{Angelopoulos2023} are promising directions for future study.

\bibliography{HSML_ref}

\end{document}